\documentclass{article}
\usepackage{ijcai26}

\usepackage{times}
\usepackage{soul}
\usepackage{url}
\usepackage[hidelinks]{hyperref}
\usepackage[utf8]{inputenc}
\usepackage[small]{caption}
\usepackage{graphicx}
\usepackage{amsmath,amssymb}
\usepackage{amsthm}
\usepackage{booktabs}
\usepackage{algorithm}
\usepackage{algorithmic}
\usepackage[switch]{lineno}

\newcommand{\dom}{\Omega}
\newcommand{\bdry}{\partial\dom}
\newcommand{\spacedim}{d}
\newcommand{\vardim}{m}
\newcommand{\timeh}{T}
\newcommand{\xx}{\mathbf{x}}
\newcommand{\Fop}{\mathcal{F}}
\newcommand{\Bop}{\mathcal{B}}
\newcommand{\Hop}{\mathcal{H}}
\newcommand{\Gop}{\mathcal{G}}
\newcommand{\spec}{a}
\newcommand{\known}{\boldsymbol{\mu}}
\newcommand{\unk}{\boldsymbol{\theta}}
\newcommand{\params}{\phi}

\newcommand{\Lphys}{\mathcal{L}_{\mathcal{F}}}
\newcommand{\Lbc}{\mathcal{L}_{\mathcal{B}}}
\newcommand{\Lic}{\mathcal{L}_{\mathrm{IC}}}
\newcommand{\Ldata}{\mathcal{L}_{\mathrm{data}}}

\title{Learning PDE Solvers with Physics and Data: A Unifying View of\\ Physics-Informed Neural Networks and Neural Operators}

\author{
	Yilong Dai$^{1}$,
	Shengyu Chen$^{2}$,
	Ziyi Wang$^{3}$,
	Xiaowei Jia$^{2}$,
	Yiqun Xie$^{3}$,
	Vipin Kumar$^{4}$,
	Runlong Yu$^{1}$\thanks{Runlong Yu is the corresponding author.}\\
	\affiliations
	$^{1}$University of Alabama,
	$^{2}$University of Pittsburgh,
	$^{3}$University of Maryland,
	$^{4}$University of Minnesota\\
	\emails
	\{ydai17, ryu5\}@ua.edu,
	\{shc160, xiaowei\}@pitt.edu,
	\{zoewang, xie\}@umd.edu,
	kumar001@umn.edu
}

\begin{document}
	\maketitle
	
	\begin{abstract}
		Partial differential equations (PDEs) are central to scientific modeling. 
		Modern workflows increasingly rely on learning-based components to support model reuse, inference, and integration across large computational processes.
		Despite the emergence of various physics-aware data-driven approaches, the field still lacks a unified perspective to uncover their relationships, limitations, and appropriate roles in scientific workflows.
		To this end, we propose a unifying perspective to place two dominant paradigms: Physics-Informed Neural Networks (PINNs) and Neural Operators (NOs), within a shared design space.
		We organize existing methods from three fundamental dimensions: what is learned, how physical structures are integrated into the learning process, and how the computational load is amortized across problem instances.
		In this way, many challenges can be best understood as consequences of these structural properties of learning PDEs.
		By analyzing advances through this unifying view, our survey aims to facilitate the development of reliable learning-based PDE solvers and catalyze a synthesis of physics and data.
	\end{abstract}
	
	\section{Introduction}
	\label{sec:intro}
	
	Partial differential equations (PDEs) are the core language of scientific modeling. They are widely used in fields such as fluid dynamics, electromagnetics, climate modeling, materials science, and biological systems.
	Although classical numerical solvers are effective for single high-fidelity simulations, many modern scientific workflows increasingly require repeated solutions to PDEs with varying parameters, boundary or initial conditions, geometries, or partial observational data.
	Such multi-query scenarios naturally arise in design optimization, uncertainty quantification (UQ), inverse problems, and data assimilation.
	In these cases, the cost of repeatedly executing traditional discretization solution processes can become a bottleneck, especially for high-dimensional, multiscale, or geometrically complex problems, and when sparse or noisy data are incorporated.
	
	These bottlenecks motivated research into \emph{learning-based PDE solvers}, methods that move from computing solutions per instance to learning reusable representations through physics, data, or both. Early purely data-driven approaches can achieve good interpolation performance, but lack generalizability, especially when geometry, boundary conditions, or regimes change. In addition, they may violate physical constraints, including conservation laws. Consequently, these limitations led to two dominant physics-aware paradigms: \emph{Physics-Informed Neural Networks (PINNs)} and \emph{Neural Operators (NOs)}.
	PINNs approximate individual solutions as neural functions, trained to satisfy governing equations and auxiliary conditions, optionally incorporating observational data~\cite{raissi2019physics}.  NOs, in contrast, learn the mapping between instance specifications and solution fields covering a family of PDEs, enabling rapid amortized inference after offline training~\cite{lu2021deeponet,li2021fourier}.
	
\begin{figure*}[t]
		\centering
		\includegraphics[width=\textwidth]{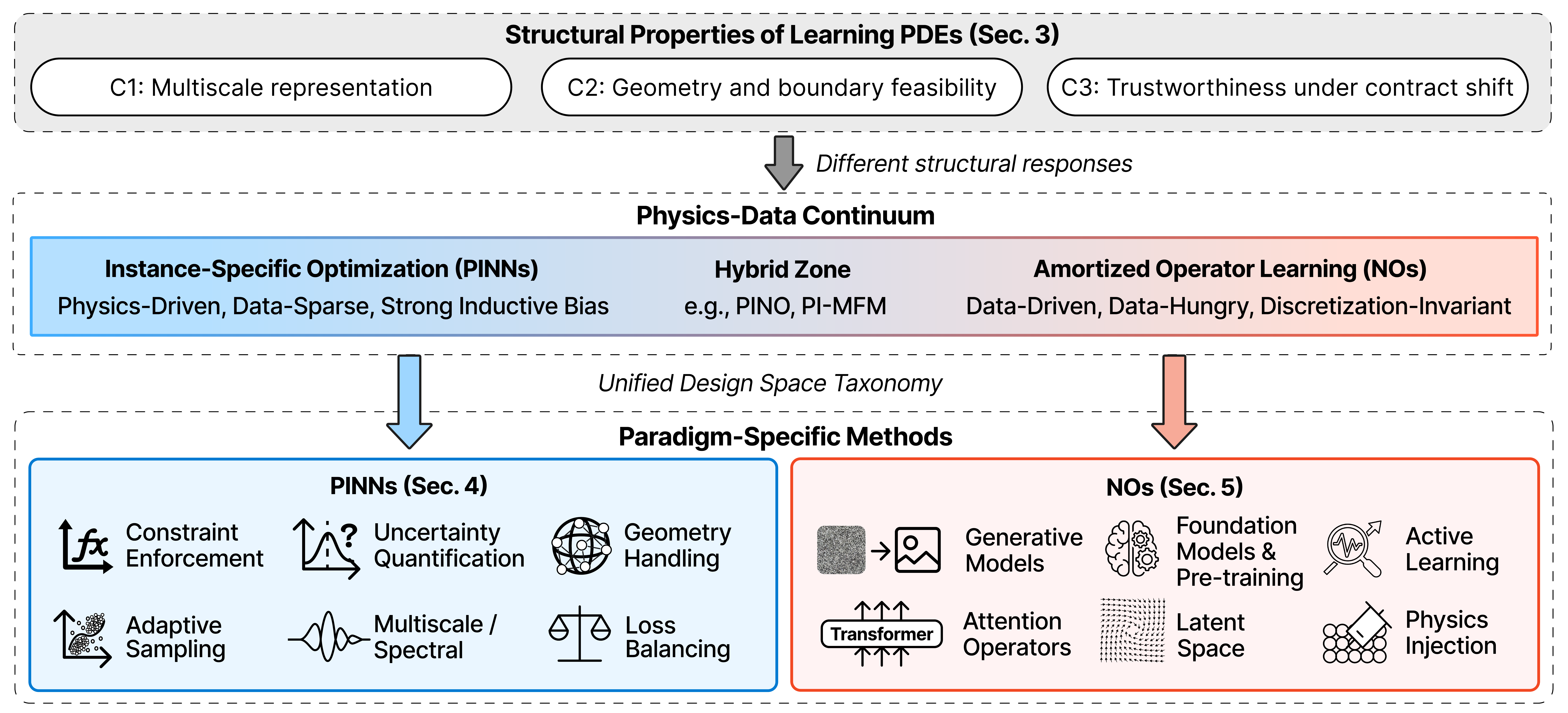}
		\caption{A unifying view of learning-based PDE solvers.
			Top: three structural properties (C1--C3) that challenge all learning 
			paradigms (Section~\ref{sec:crosscut}). Middle: a continuum shaped 
			by what is learned and how physics enters supervision, organized within 
			a unified design space taxonomy. Bottom: paradigm-specific innovations 
			for PINNs (left, Section~\ref{sec:pinn}) and neural operators (right, 
			Section~\ref{sec:no}).}
		\label{fig:overview}
	\end{figure*}

	Typically, the community views PINNs and NOs as competing methods for learning-based PDE solving. In this survey, however, we take a different perspective.
	We view these two paradigms as complementary methods within a shared \emph{physics--data continuum} and a common design space shaped by workflow requirements (see Figure~\ref{fig:overview}).
	From this perspective, PINNs and NOs have a common objective: approximating the mapping from PDE specifications to their solutions.
	Therefore, both paradigms face inherent constraints imposed by PDE solution families, including multi-scale representation, geometric and boundary feasibility, and trustworthiness under contract shift.
	Meanwhile, the two paradigms also have systematic differences, because they stem from their fundamental structural decisions:
	what is being learned (an instance-specific solution versus a family-level operator),
	how physical structure is incorporated into learning (physics-as-supervision versus paired supervision with physics injection),
	and where computational cost is amortized (instance-by-instance optimization versus amortized inference across instances).
	These decisions lead to unique regimes of difficulty and failure patterns for each.

	\paragraph{Scope and contributions.}
	This survey comprehensively reviews the literature from 2017 to 2026 (primarily from top AI and machine learning conferences as well as mathematics and physics journals). It covers the emergence and recent development of PINNs, NOs, and hybrid variants.
	Our contributions have three aspects:
	(i) a unifying taxonomy based on structural design choices and workflow contracts;
	(ii) a cross-paradigm analysis to examine structural properties of learning PDEs; and
	(iii) application-oriented method choice, as well as open challenges and opportunities towards scientific deployment.
	
	\paragraph{Organization.}
	Section~\ref{sec:designspace} defines a general PDE formulation and the unified design space.
	Following this, Section~\ref{sec:crosscut} characterizes core structural properties shared across two learning paradigms.
	Next, Sections~\ref{sec:pinn} and~\ref{sec:no} separately review paradigm-specific developments of PINNs and NOs.
	After that, Section~\ref{sec:applications} discusses applications and method choice.
	Then, Section~\ref{sec:challenges} outlines open challenges and future opportunities.
	Finally, Section~\ref{sec:conclusion} concludes the survey.
	
	\section{Problem Setup and Design Space}
	\label{sec:designspace}
	
	We introduce a consistent problem setup for learning-based PDE solvers.
	Our goal is to clarify what is being solved, what is learned, and how learning formulations relate within a shared design space.
	
	\paragraph{PDE problems.}
	We consider PDEs defined on a spatial domain $\dom \subset \mathbb{R}^{\spacedim}$ and, when applicable, a time interval $t \in [0,\timeh]$.
	The unknown is a field
	\begin{equation}
		u:\dom \times [0,\timeh] \rightarrow \mathbb{R}^{\vardim},
		\label{eq:field_def}
	\end{equation}
	with $\spacedim$ spatial dimensions and $\vardim$ physical variables.
	Many PDEs can be written abstractly as
	\begin{equation}
		\Fop\!\left(
		u(\xx,t), \nabla u(\xx,t), \ldots, D^k u(\xx,t);
		\xx, t; \known, \unk
		\right)=0,
		\label{eq:general_pde}
	\end{equation}
	where $\Fop$ denotes the governing differential operator;
	$\known$ collects \emph{known} instance specifications (e.g., coefficients, forcing, geometry descriptors, boundary/initial data);
	and $\unk$ denotes \emph{unknown} quantities to be inferred in inverse settings.
	
	For time-dependent problems, the PDE is complemented by initial and boundary conditions
	\begin{equation}
		u(\xx,0)=u_0(\xx),\qquad \Bop[u](\xx,t)=g(\xx,t)\quad \text{on }\; \bdry,
		\label{eq:bc_ic}
	\end{equation}
	where $\Bop$ is a boundary operator and $\bdry$ is the boundary of $\dom$.
	
	\paragraph{Well-posedness and learning.}
	Classical analysis studies whether Eq.~\eqref{eq:general_pde}--\eqref{eq:bc_ic} is well-posed.
	In learning-based settings, these conditions are often weakened or effectively violated under sparse/noisy observations, unknown parameters, or out-of-contract deployment.
	Stability is then promoted through inductive bias, regularization, and diagnostics rather than guaranteed purely by the PDE.
	These departures from classical well-posedness are precisely what motivate learning-based formulations under partial observation and task-specific workflow regimes.
	
	\paragraph{Observations.}
	When data are available, they are often partial and noisy:
	\begin{equation}
		y_i=\Hop[u](\xx_i,t_i)+\varepsilon_i,
		\label{eq:obs}
	\end{equation}
	where $\Hop$ is an observation operator and $\varepsilon_i$ denotes noise.
	
	\paragraph{Workflow regimes.}
	We will refer to:
	\emph{forward solving} (given $\known$, infer $u$),
	\emph{inverse/identification} (parts of $\unk$ unknown, infer $u$ and $\unk$ from $y$),
	and \emph{data assimilation} (recover spatiotemporal state under sparse/noisy measurements with stability requirements).
	
	\paragraph{Solution learning (PINN baseline).}
	\label{sec:solution_learning_baseline}
	
	Solution-learning approaches represent the unknown field by $u_{\params}(\xx,t)$ (typically a coordinate network).
	Physics enters via residuals, e.g.,
	\begin{equation}
		r_{\Fop}(\xx,t)=\Fop\!\left(
		u_{\params}(\xx,t), \nabla u_{\params}(\xx,t), \ldots; \xx, t; \known, \unk
		\right),
		\label{eq:residual}
	\end{equation}
	with analogous residuals for Eq.~\eqref{eq:bc_ic}.
	A canonical formulation minimizes a composite objective
	\begin{equation}
		\min_{\params,\unk}~
		\lambda_{\Fop}\Lphys
		+\lambda_{\Bop}\Lbc
		+\lambda_{\mathrm{IC}}\Lic
		+\lambda_{\mathrm{data}}\Ldata,
		\label{eq:composite_loss}
	\end{equation}
	where $\Lphys$ measures the PDE residual in Eq.~\eqref{eq:residual},
	$\Lbc$ and $\Lic$ measure boundary and initial-condition violations in Eq.~\eqref{eq:bc_ic},
	and $\Ldata$ measures the mismatch to observations through $\Hop$ (Eq.~\eqref{eq:obs});
	each term is typically an average squared residual (or misfit) over sampled collocation, boundary/initial, and measurement points,
	and the weights $\lambda$ control their relative importance.
	This baseline corresponds to the core PINN template~\cite{raissi2019physics}:
	learn a \emph{single instance} by treating physics (and optional observations) as supervision.
	
	\paragraph{Operator learning (NO baseline).}
	\label{sec:operator_learning_baseline}
	
	Operator learning targets a family of PDE instances.
	Let $\spec$ denote an \emph{instance specification} encoding $\known$ as one or more input functions.
	An operator is a map
	\begin{equation}
		u = \Gop(\spec),
		\label{eq:operator_map}
	\end{equation}
	trained on paired data $\{(\spec_j,u_j)\}_{j=1}^N$.
	Canonical realizations include DeepONet-style factorization~\cite{lu2021deeponet} and Fourier Neural Operators (FNO)~\cite{li2021fourier}.
	A defining deployment goal is often \emph{discretization-robust inference}:
	evaluating $\Gop$ on discretizations (and sometimes geometries) not identical to those used in training.
	
	\paragraph{A unifying taxonomy.}
	\label{sec:taxonomy}
	
	We organize the field through three structural choices.
	
	\paragraph{(T1) Object of learning.}
	PINNs learn an instance-specific solution $u_{\params}(\xx,t)$.
	Neural operators learn a family-level operator $\Gop$ mapping specifications $\spec$ to solutions.
	
	\paragraph{(T2) Supervision interface.}
	PINNs treat physics constraints as primary supervision.
	Neural operators treat paired data as primary supervision; physics is injected through architecture, regularization, refinement, or diagnostics~\cite{shu2023physics,bastek2025physicsinformed}.
	
	\paragraph{(T3) Amortization locus.}
	PINNs amortize \emph{supervision} but typically require per-instance training.
	Neural operators amortize \emph{inference compute} but require data generation and offline training.
	
	\paragraph{Workflow contracts.}
	These structural choices are induced by the \emph{workflow contract} in which a solver is embedded, and they directly shape the dominant failure modes encountered.
	For PINNs, physics-as-supervision combined with per-instance optimization places numerical conditioning and stiff multi-objective training at the center of the workflow.
	For neural operators, paired supervision with amortized inference shifts the risk toward distribution shift, discretization mismatch, and generalization of geometry at deployment.
	
	More generally, a learning-based PDE solver can be interpreted relative to a \emph{workflow contract}, which specifies the perturbations permitted at deployment, the invariants that must be preserved, and the operational signals required   (e.g., latency, reliability, and diagnostics).
	In scientific computing pipelines, such contracts are typically characterized along a small set of recurring axes:
	(i) \emph{instance variability}, including changes in coefficients, forcing terms, boundary or initial conditions, and unknown parameters $\unk$ across queries;
	(ii) \emph{geometry and discretization variation}, encompassing changes in domain shape, boundary configuration, and grid or mesh representations between training and evaluation;
	(iii) \emph{temporal horizon}, ranging from single-step inference to long-horizon rollout where small biases can accumulate over time;
	(iv) \emph{available supervision}, spanning paired simulation data, sparse or noisy observations via $\Hop$ (Eq.~\eqref{eq:obs}), and physics constraints alone; and
	(v) \emph{deployment validity and trust signals}, such as boundary feasibility, conservation properties, uncertainty estimates, out-of-distribution detection, and fallback or refinement mechanisms.
	
	Section~\ref{sec:crosscut} analyzes structural properties of PDE solution families that make satisfying these workflow contracts challenging in practice, while Sections~\ref{sec:pinn} and~\ref{sec:no} examine how different learning paradigms expose distinct interfaces for meeting a given contract.
	
	\section{Structural Properties of Learning PDEs}
	\label{sec:crosscut}
	
	This section is deliberately paradigm-agnostic. Rather than enumerating algorithms, we characterize shared \emph{structural properties} of PDE solution families that act as persistent pressure points across learning-based workflows.
	These properties explain \emph{why} learning PDEs is difficult and \emph{which evaluation signals} are necessary to assess reliability beyond average accuracy.
	Paradigm-specific mechanisms and design choices are reviewed in Sections~\ref{sec:pinn} and~\ref{sec:no}.
	We focus on three recurring properties:
	\emph{(C1) multiscale representation},
	\emph{(C2) geometry and boundary feasibility},
	and \emph{(C3) trustworthiness under workflow contract shift}.
	
	\subsection{C1: Multiscale representation}
	\label{sec:multiscale}
	
	PDE solutions are inherently multiscale: sharp boundary layers, intermittent structures, and broadband spectra can coexist with smooth regions.
	Multiscale difficulty is not solely a matter of expressivity; it arises from the interaction between (i) available learning signals, (ii) finite representations or basis choices, and (iii) the deployment horizon over which predictions are evaluated.
	
	\paragraph{Uneven learnability across scales.}
	Common learning signals (such as residuals, paired supervision, or partial observations) typically allocate error unevenly across frequencies.
	Low-frequency components are often learned first, while high-frequency content may be delayed, smeared, or aliased depending on representation and discretization.
	
	\paragraph{Pathologies of differential and nonlocal priors.}
	Strong-form differential supervision can amplify high-frequency noise through derivatives, while operator architectures relying on global mixing may blur localized structure when fine-scale content is poorly represented.
	In long-horizon settings, small scale-dependent biases can accumulate and alter qualitative behavior.
	
	Beyond global MSE, multiscale performance should be evaluated using signals aligned with the workflow contract, including:
	(i) spectral or band-limited error metrics,
	(ii) localized errors near interfaces or boundaries,
	(iii) error growth under rollout (one-step versus multi-step),
	and (iv) task-level invariants sensitive to fine-scale structure when relevant.
	
	\subsection{C2: Geometry and boundary feasibility}
	\label{sec:geometry}
	
	Geometry and boundary conditions define the feasible solution set through the domain $\dom$, boundary $\bdry$, and boundary operator $\Bop$ (Eq.~\eqref{eq:bc_ic}).
	Under deployment perturbations, geometry and boundary variation are among the most common sources of workflow contract violation.
	
	\paragraph{Feasibility is an interface problem.}
	A solver must specify how geometric and boundary information enters computation.
	This interface determines whether feasibility is enforced by construction, weakly encouraged, or left implicit.
	Concretely, this requires consistent choices for:
	(i) \emph{domain representation} (grid, mesh, graph, or manifold coordinates),
	(ii) \emph{boundary representation} (boundary identification and prescribed conditions),
	and (iii) \emph{feasibility enforcement} (how constraints are imposed or encoded).
	
	\paragraph{Mismatch can be silent.}
	When geometry or boundary information is under-specified or inconsistently represented, violations may not appear in the average training error.
	Instead, they surface abruptly under deployment perturbations, such as new domain shapes, boundary types, or discretizations. 
	
	Geometry and boundary feasibility should be evaluated using:
	(i) boundary violation metrics (e.g., Dirichlet or Neumann mismatch),
	(ii) conservation or compatibility metrics tied to boundaries when relevant,
	(iii) sensitivity to geometry perturbations,
	and (iv) stability under boundary variation.
	
	\subsection{C3: Trustworthiness under contract shift}
	\label{sec:trust}
	
	Scientific deployment requires deciding when predictions remain trustworthy under shifts in the workflow contract.
	Mean accuracy alone is insufficient, as failures can be silent and operationally costly.
	Trustworthiness is therefore a system-level property, encompassing both predictions and the mechanisms used to assess their validity.
	
	\paragraph{Distribution shift is structured.}
	Contract shift typically occurs along a small number of structured axes, including parameters or regimes, geometry or boundary types, discretization or resolution, rollout horizon, and observation operators.
	Methods that perform well under interpolation may fail under these shifts without explicit warning signals.
	
	\paragraph{Operational signals and fallback mechanisms.}
	Trustworthy systems surface uncertainty, detect out-of-contract inputs, and couple predictions to validity checks, such as residual or feasibility diagnostics, that can trigger refinement or fallback.
	
	Trustworthiness evaluation should include:
	(i) calibration and sharpness when probabilistic outputs are available,
	(ii) out-of-distribution detection and selective prediction,
	(iii) constraint violation rates and stability indicators,
	and (iv) the ability to trigger refinement or fallback under stress tests.
	These signals should be aligned with the workflow contract, not only average error.
	
	
	\section{Physics-Informed Neural Networks}
	\label{sec:pinn}
	
	This section reviews developments implied by the PINN structural choice:
	learning an instance-specific solution via physics-based supervision and per-instance optimization (Eq.~\eqref{eq:composite_loss}).
	The shared pressures (\emph{why} learning PDEs is hard) and the recommended evaluation signals are defined once in Section~\ref{sec:crosscut}.
	Here we focus strictly on the \emph{PINN interfaces}---residual construction, constraint enforcement, representation, optimization, and sampling---that determine whether a PINN meets a given workflow contract.
	
	%
	\subsection{Baseline recap and canonical failure modes}
	\label{sec:pinn_baseline_fail}
	
	Physics-Informed Neural Networks, popularized by Raissi et al.~\cite{raissi2019physics}, train an instance-specific neural solution by minimizing a weighted sum of physics residuals, auxiliary-condition violations, and optional data mismatch (Eq.~\eqref{eq:composite_loss}).
	Their appeal is data efficiency, mesh-free evaluation, and natural support for inverse problems and data assimilation (unknown $\unk$ and partial observations in Eq.~\eqref{eq:obs}).
	
	Empirically, the PINN research agenda is shaped by training-loop failure modes:
	(i) strong-form residuals can be poorly conditioned for stiff PDEs and long-time dynamics;
	(ii) soft penalties can converge to low average loss but leave localized feasibility violations of Eq.~\eqref{eq:bc_ic};
	(iii) multi-term objectives exhibit gradient conflicts across $\Lphys,\Lbc,\Lic,\Ldata$;
	(iv) uniform collocation under-allocates samples to localized complexity;
	(v) automatic differentiation makes residual evaluation expensive for high-order derivatives and large collocation sets.
	Most PINN variants can be interpreted as design knobs targeting these mechanisms.
	
	\subsection{Design knobs for practical PINNs}
	\label{sec:pinn_knobs}
	
	We group the literature into five knobs that map directly to the terms in Eq.~\eqref{eq:composite_loss} and the failure modes above.
	
	\subsubsection{(K1) Residuals: strong vs.\ weak}
	Pointwise strong-form residuals are simple but can be stiff or noisy.
	Weak/variational/integral objectives improve conditioning by aligning the learning signal with Galerkin/FEM reasoning~\cite{saleh2024learningintegrallossesphysics,Gao_2022}.
	
	\subsubsection{(K2) Constraints: soft vs.\ hard}
	To meet feasibility requirements under geometry/boundary variation (contract axis C2), many PINNs move beyond soft penalties.
	Hard/semi-hard satisfaction can be achieved by constraint-satisfying parameterizations, projection, or physics-embedded discretized operators~\cite{chen2025pinpphysicsinformedneuralpredictor,CHEN2021110624,chalapathi2024scalingphysicsinformedhardconstraints}.
	These designs directly target boundary and auxiliary operators $\Bop$ in Eq.~\eqref{eq:bc_ic}.
	
	\subsubsection{(K3) Representation: spectral and multiscale}
	Rather than re-motivating multiscale difficulty (Section~\ref{sec:multiscale}), we focus on PINN-side remedies that modify the \emph{training interface}:
	representations that reduce derivative amplification in Eq.~\eqref{eq:residual} and accelerate learning of high frequencies.
	This includes Fourier encodings and multiresolution architectures~\cite{cooley2025fourierpinnsstrongboundary,musgrave2024fourierdomainphysicsinformed,han2024ugridefficientandrigorousneuralmultigrid,kang2025pigphysicsinformedgaussiansadaptive,fang2024solvinghighfrequencymultiscale}.
	
	\subsubsection{(K4) Optimization: balance and curvature}
	PINN training behaves like a numerical procedure over a discretized objective:
	sampling, resolution, and loss weighting act like discretization and step-size choices.
	Dynamic balancing mitigates gradient conflicts~\cite{Xiang_2022,Bischof_2025,hwang2025dualconegradientdescent,liu2025configconflictfreetrainingphysics}.
	Curvature-aware and preconditioned strategies stabilize stiff dynamics and long-time integration~\cite{chen2024tengtimeevolvingnaturalgradient,schwencke2025anagramnaturalgradientrelative,dangel2024kroneckerfactoredapproximatecurvaturephysicsinformed,finzi2023stablescalablemethodsolving}.
	Failure-mode analyses relate difficulties to loss landscape structure~\cite{bonfanti2024challengesnonlinearregimephysicsinformed,rathore2024challengestrainingpinnsloss,AHMADIDARYAKENARI2026111393}.
	Meta-learned optimization transfers update rules across instances~\cite{boudec2025learningneuralsolverparametric,wandel2025metamizerversatileneuraloptimizer}.
	
	\subsubsection{(K5) Sampling: adaptive collocation}
	Collocation is the discretization of the training signal.
	Residual-based refinement and physics-aware sampling allocate collocation where needed~\cite{Wu_2023,guo2022noveladaptivecausalsampling}.
	Information acquisition views collocation as experimental design~\cite{lau2024pinnaclepinnadaptivecollocation,wu2024ropinnregionoptimizedphysicsinformed,hemachandra2025piedphysicsinformedexperimentaldesign,tang2024adversarialadaptivesamplingunify}.
	
	\subsection{Scaling and geometry for PINNs}
	\label{sec:pinn_geometry}
	
	We detail the \emph{PINN-side geometry interface}: how domain and boundary information enters the \emph{training loop} through residual evaluation, constraint enforcement, and sampling.
	
	\paragraph{Residual evaluation on discretizations.}
	On meshes, point clouds, or manifolds, the central question is how to evaluate the operator signal consistently.
	This includes discretized differential operators, weak forms on elements/patches, intrinsic coordinates on manifolds, and geometry-aware quadrature rules.
	Graph-based PINNs and message passing can be viewed as transporting local operator structure onto discrete neighborhoods~\cite{zeng2025phympgnphysicsencodedmessagepassing,garnier2025meshmaskphysicsbasedsimulationsmasked,yu2025piorfphysicsinformedollivierricciflow,jiang2023phygnnetsolvingspatiotemporalpdes,jia2023physics}.
	
	\paragraph{Boundary and constraint enforcement.}
	Geometry enters PINNs not only through $\Fop$ but also through the practical representation of $\bdry$ and the operator $\Bop$.
	Hard/semi-hard approaches build feasibility into parameterizations or enforce it via projection, reducing reliance on sampling density near boundaries~\cite{chen2025pinpphysicsinformedneuralpredictor,CHEN2021110624,chalapathi2024scalingphysicsinformedhardconstraints}.
	
	\paragraph{Sampling on geometry.}
	Adaptive strategies must respect geometry: boundary sampling, curvature-aware sampling on manifolds, and refinement near interfaces where feasibility violations concentrate.
	Manifold-aware PINNs use intrinsic coordinates and curvature-aware designs to reduce coordinate-induced artifacts~\cite{SAHLICOSTABAL2024107324,li2024finite,zhou2025ltpinnlagrangiantopologyconsciousphysicsinformed,liao2025curvatureaware}.
	
	\subsection{Computational efficiency}
	\label{sec:pinn_efficiency}
	
	A central bottleneck of PINNs is repeated residual evaluation with automatic differentiation.
	Strategies include domain decomposition and parallelization~\cite{cho2023separablephysicsinformedneuralnetworks,moseley2021scaling,shukla2021parallelphysicsinformedneuralnetworks,Moseley_2023,WU2022111588},
	hybrid AD and numerical differentiation~\cite{sharma2023acceleratedtrainingphysicsinformedneural,Chiu_2022},
	and hybrid solver/operator-splitting designs~\cite{9403414,nguyen2024parcv2physicsawarerecurrentconvolutional,CHEN2024112846}.
	
	\subsection{Inverse problems, uncertainty, and diagnostics}
	\label{sec:pinn_uq}
	
	For PINNs, trustworthiness (Section~\ref{sec:trust}) is largely mediated by the \emph{inference interface}:
	unknowns $\unk$ and sparse observations can render learning ill-posed, so reliable deployment requires uncertainty estimates and diagnostics tied to the contract.
	Bayesian/variational methods and robust regression variants quantify uncertainty under sparse/noisy data~\cite{Linka_2022,YANG2021109913,rojas2024robustvariationalphysicsinformedneural,hamelijnck2025physicsinformedvariationalstatespacegaussian,peng2022robustregressionhighlycorrupted}.
	Generative formulations provide distributional field inference~\cite{Daw_2021,Gao_2022,bastek2025physicsinformeddiffusionmodels}.
	Benchmarks and certification support diagnosis and reproducibility~\cite{hao2023pinnaclecomprehensivebenchmarkphysicsinformed,eiras2024efficienterrorcertificationphysicsinformed,Basir_2023}.
	
	\subsection{Bridging toward reuse}
	\label{sec:pinn_bridge}
	
	Although vanilla PINNs are per-instance solvers, hypernetworks and parameter-conditioned PINNs move toward reuse across instances~\cite{belbuteperes2021hyperpinnlearningparameterizeddifferential,cho2024parameterizedphysicsinformedneuralnetworks},
	and meta-learning amortizes adaptation across PDE families~\cite{qin2022metapdelearningsolvepdes,PENWARDEN2023111912,Psaros_2022,toloubidokhti2024dats,chen2023physics}.
	These directions connect naturally to hybrids by shifting part of per-instance optimization into reusable representations.
	

	\section{Neural Operators}
	\label{sec:no}
	
	This section reviews developments implied by the operator-learning structural choice:
	learning an instance-to-solution mapping with paired data (Eq.~\eqref{eq:operator_map}) for amortized inference and discretization-robust deployment.
	The shared pressures and evaluation signals are defined once in Section~\ref{sec:crosscut}.
	Here we focus strictly on \emph{operator interfaces}: how $\spec$ and $u$ are represented, how nonlocal mixing is implemented, and how robustness is engineered under the workflow contract.
	
	
	\subsection{Baseline recap and dominant risks}
	\label{sec:no_baseline_risks}
	
	A neural operator learns a map $u=\Gop(\spec)$ from an input function space to a solution function space (Eq.~\eqref{eq:operator_map}), trained on paired instances.
	A defining goal is to evaluate $\Gop$ on discretizations different from those used in training, shaping how inputs/outputs are parameterized and how nonlocal mixing is implemented.
	Canonical templates include DeepONet~\cite{lu2021deeponet} and FNO~\cite{li2021fourier}, but most later operators can be interpreted as modifying:
	(i) function parameterization (how $\spec$ and $u$ are discretized or encoded),
	(ii) nonlocal interaction mechanisms (how information mixes globally/locally),
	and (iii) geometry/boundary interfaces (how domains and boundary data enter $\spec$).
	
	The dominant risks differ from PINNs because operators are deployed as surrogates:
	(i) \emph{dataset coverage} and distribution shift in $\spec$,
	(ii) \emph{discretization and geometry transfer} (contract axis C2),
	and (iii) \emph{long-horizon error accumulation} in dynamical rollouts (C3).
	Accordingly, much of the literature focuses on representation and interfaces robust under contract transfer, and on system-level mechanisms (diagnostics/refinement/fallback) for safe deployment.
	
	\subsection{Geometry, discretization, boundary interfaces}
	\label{sec:no_geometry}
	
	We detail the \emph{operator-side geometry interface}: how to encode $\spec$ and decode $u$ so that $u=\Gop(\spec)$ remains well-defined across grids/meshes/domains, and how to inject boundary information without breaking discretization robustness.
	
	\paragraph{Grid vs mesh/graph operators as interface choices.}
	Grid-based operators (e.g., FNO-style) assume a structured discretization for efficient global mixing.
	Mesh/graph operators generalize mixing to irregular discretizations and can encode conservation and symmetry through message passing~\cite{wang2024beno,zeng2025phympgn,horie2024fluxgnn,xu2024equivariant}.
	This is to make the \emph{mapping} stable when discretizations change.
	
	\paragraph{Domain variation and coordinate alignment.}
	When domains vary, operators often introduce alignment mechanisms:
	reference mappings/diffeomorphisms, intrinsic bases on manifolds, or kernel/integral approximations that operate in coordinate-free ways~\cite{chen2024learning,zhao2025diffeomorphism,quackenbush2024geometricneuraloperatorsgnps,lingsch2024beyond,Di2026FGConvNO}.
	Adaptive discretization complements model capacity with classical meshing choices~\cite{hu2024better}.
	
	\paragraph{Boundary injection.}
	Boundary conditions can enter $\spec$ as input channels, as separate boundary fields, or via decompositions that separate boundary-driven and interior effects.
	Analytical constrained representations encode boundary satisfaction by construction~\cite{dalton2024bcgp}, while boundary-aware decompositions (e.g., Green's-function-inspired) separate boundary and interior contributions~\cite{wang2024beno}.
	
	\subsection{Multiscale and high-frequency structure}
	\label{sec:no_multiscale}
	
	We focus on \emph{operator-side multiscale interfaces}: how global mixing and basis choices handle (or fail to handle) fine-scale content under resolution and distribution shifts.
	
	\paragraph{Spectral truncation and aliasing.}
	Fourier-style mixing is efficient for smooth fields but truncation can smooth localized high-frequency features, and aliasing can dominate error under resolution shifts.
	Spectral operator variants extend FNO via factorization and constrained mixing to reduce sensitivity~\cite{tran2023factorized,kossaifi2024multigrid,xiao2024amortized,behroozi2025sensitivityconstrained}.
	
	\paragraph{Locality and scale separation.}
	Wavelet/multiwavelet and multigrid-inspired operators introduce locality and explicit scale separation to better capture sharp features and localized dynamics~\cite{hu2025wavelet,xiao2023coupled,he2024mgno,luo2024hierarchical}. Frequency-adaptive spectral losses offer an orthogonal, loss-side route to multi-scale fidelity~\cite{dai2026pest}.
	On non-Euclidean domains, spectral bases generalize beyond Fourier (e.g., spherical harmonics)~\cite{bonev2023sphericalfourierneuraloperators}.
	
	\subsection{Latent and attention-based operators}
	\label{sec:no_latent}
	
	Latent token operators compress inputs into fixed-size latent states for scalable mixing~\cite{wang2024latent,alkin2024upt}.
	Attention can be interpreted as learned nonlocal kernels~\cite{yu2024nao}.
	Cross-attention enables few-shot operator learning by conditioning on context pairs~\cite{Jiao2025}.
	State-space models offer linear-complexity alternatives for long sequences~\cite{HU2026108496}, and compressed generative pipelines enable conditional generation and super-resolution~\cite{zhou2024text2pde}.
	
	\subsection{Physics injection}
	\label{sec:no_physics}
	
	Physics can be incorporated via architecture, objectives, or correction steps to improve robustness under shift.
	This is an operator-side \emph{augmentation interface}: physics is not the primary supervision (as in PINNs), but an added mechanism for validity and transfer.
	Importantly, these mechanisms also define the practical \emph{hybrid (PI-operator)} regime: an operator provides fast, amortized \emph{proposals} (Eq.~\eqref{eq:operator_map}), while physics constraints provide \emph{correction} and \emph{validation} signals (e.g., residuals and boundary operators, Eq.~\eqref{eq:residual}--\eqref{eq:bc_ic}), optionally triggering lightweight per-instance refinement.
	Representative examples include Physics-Informed Neural Operators (PINO)~\cite{li2024physics}, as well as residual-regularized and refinement-based operator pipelines.
	
	\paragraph{Conservation and symmetry.}
	Concretely, flux-form message passing enforces conservation via antisymmetric exchange~\cite{horie2024fluxgnn}.
	Equivariant constructions preserve symmetries and stabilize transfer~\cite{xu2024equivariant,shumaylov2025lie,Liu2025NeuralMD}.
	
	\paragraph{Boundary handling.}
	Similarly, boundary-aware decompositions separate boundary-driven and interior effects~\cite{wang2024beno}.
	Analytical constrained representations encode boundary satisfaction by construction~\cite{dalton2024bcgp}.
	Physics-informed DeepONet variants add boundary channels for improved enforcement~\cite{cho2025physicsinformed}.
	
	\paragraph{Residual-regularized operators and refinement.}
	Residual penalties and physics-informed refinement connect operator learning to the hybrid zone: the operator proposes; physics corrects and certifies, and refinement can be invoked when diagnostics indicate out-of-contract behavior.

	\subsection{Data efficiency, robustness, system-level cost}
	\label{sec:no_data}
	
	Active learning selects informative PDE instances to reduce data-generation costs~\cite{musekamp2025active}.
	Multifidelity learning combines cheap low-fidelity simulations with scarce high-fidelity data~\cite{Velikorodny2025}.
	Under distribution shift, uncertainty-aware operators improve calibration and OOD detection~\cite{mouli2024diverseno}.
	End-to-end efficiency includes data generation, recycling/warm-start during corpus construction~\cite{wang2024accelerating}, and memory-stable/mixed-precision implementations~\cite{tu2024guaranteed}.
	
	\subsection{Probabilistic and generative neural operators}
	\label{sec:no_prob}
	
	Probabilistic operators using proper scoring rules yield calibrated function-space uncertainty~\cite{bulte2025probabilistic}.
	Diffusion-based refinement recovers fine-scale structure lost under truncation or coarse representations, supports iterative error correction in long rollouts, and can incorporate physics residuals during generation~\cite{NEURIPS2023_d529b943,jiang2025integrating,liu2025difffno,oommen2025integratingneuraloperatorsdiffusion,bastek2025physicsinformed,shu2023physics}.
	Generative operator pipelines support inverse design and joint inference from partial observations~\cite{wu2024compositional,huang2024diffusionpde}.
	
	\subsection{Foundation models for PDE operators}
	\label{sec:no_foundation}
	
	At a larger scale, operator foundation models aim to transfer representations across multiple physics domains~\cite{hao2024dpot,alkin2024upt}.
	Physics-guided fine-tuning in spectral norms improves high-frequency correctness under transfer~\cite{cao2025spectralrefiner}.
	Physics-informed fine-tuning enables adaptation via residual-based objectives assembled from symbolic PDE expressions~\cite{zhu2025pimfmphysicsinformedmultimodalfoundation,ZHANG2026114537}.
	Modulation or augmentation mechanisms reduce fine-tuning cost under resolution and domain shifts~\cite{jeon2024pacfno,wang2025gridmix}.

	\section{Applications and Method Choice}
	\label{sec:applications}
	
	This section answers a practical question: \emph{when should one use PINNs, NOs, or hybrids?}
	Method choice is mainly about supervision availability, amortization payoff, and the required generalization contract under concrete workflow constraints.
	
	\paragraph{Decision axes.}
	We highlight five axes:
	(i) availability of paired simulation data versus only sparse/indirect observations (Eq.~\eqref{eq:obs});
	(ii) forward surrogate vs.\ inverse/assimilation objectives (unknown $\unk$ in Eq.~\eqref{eq:general_pde});
	(iii) number of queries over a fixed PDE family (amortization payoff for Eq.~\eqref{eq:operator_map});
	(iv) geometry/boundary complexity and how much it varies at deployment (Eq.~\eqref{eq:bc_ic});
	(v) deployment requirements (latency, differentiability, robustness, uncertainty).
	
	\paragraph{A recurring pattern.}
	When simulation coverage exists and many-query evaluation is central (design loops, UQ, digital twins), NOs are often preferred as fast surrogates.
	When data are sparse, unknown parameters must be inferred, or feasibility under constraints is paramount, PINNs provide a strong physics prior.
	Hybrids increasingly implement proposal--correct--validate with uncertainty-aware diagnostics: an operator proposes (Eq.~\eqref{eq:operator_map}), physics refines and validates (Eq.~\eqref{eq:residual}).
	
	\paragraph{Domain evidence.}
	\emph{Fluid dynamics} often involves many-query workloads in design and UQ, favoring operator surrogates; geometry-aware variants enable prediction on complex surfaces such as propeller blades~\cite{alkin2025abuptscalingneuralcfd,Catalani_2024,peyvan2025fusiondeeponetdataefficientneuraloperator,SUN2020112732,MAO2020112789,JIN2021109951,Di2026FGConvNO,dai2026pest}. PINNs/hybrids are frequently used for sensor-limited, boundary-sensitive, or regime-shifted inference.
	
	\emph{Solid mechanics and materials} often split similarly: operators amortize repeated finite-element solves across shape/loading variation, while inverse identification of latent parameters, defects, or damage states favors physics-regularized inference; history dependence motivates hybrids~\cite{HE2024117130,park2024pointdeeponetdeepoperatornetwork,CHEN2026105063,doi:10.1126/sciadv.abk0644,ZHOU2023107234,Goswami_2022,CHEN2024112846}.
	
	\emph{Geophysics and climate} emphasize non-Euclidean geometry, scale separation, sparse observations, and long rollouts. Geometry-aware operators enable fast ensemble surrogates, while PINNs address sensing-limited inverse problems. Long-horizon stability motivates conservative hybrids~\cite{bonev2023sphericalfourierneuraloperators,huSphericalMultigridNeural2025,diab2023udeeponetunetenhanceddeep,agata2024physicsinformeddeeplearningquantifies,ren2022seismicnetphysicsinformedneuralnetworks,verma2024climodeclimateweatherforecasting,shumaylov2025lie,jia2023physics,chen2023physics}.
	
	\emph{Energy systems} foreground deployment constraints: low-latency monitoring/dispatch/control favors surrogates, while inverse state estimation favors physics-regularized inference; constraint-aware hybrids maintain feasibility under shift~\cite{gopakumar2023fourierneuraloperatorplasma,varbella2024physicsinformedgnnnonlinearconstrained,9858911,WANG2024235271,wang2024physicsinformed,Nellikkath_2022}.
	
	\emph{Biomedical and molecular applications} combine expensive physics with scarce data: operator learning accelerates screening, while physics-informed and probabilistic variants stabilize ill-posed inference where uncertainty is intrinsic~\cite{kim2024gaussianplanewaveneuraloperator,10.1093/bioinformatics/btaf321,balcerak2024physicsregularizedmultimodalimageassimilation,pokkunuru2023improved,zhang2024cryogemphysicsinformedgenerativecryoelectron}. Multifidelity operators fuse simulations with in-vitro measurements in hemodynamics~\cite{Velikorodny2025}, while equivariant neural dynamics accelerate molecular modeling~\cite{Liu2025NeuralMD}.
	
	\emph{Engineering design and control} embed PDE models inside decision loops: fast differentiable surrogates favor operators; PINNs/hybrids support monitoring and constrained inference under sparse sensing where feasibility matters more than average error~\cite{matada2025generalizablemotionplanningoperator,cao2023deepneuraloperatorspredict,zhu2024pacepacingoperatorlearning,10.1111/mice.13436,uhrich2024physicsinformed,Antonelo_2024,GU2024104569}.
	
	\section{Open Challenges and Future Opportunities}
	\label{sec:challenges}
	
	Method choice clarifies what each paradigm can deliver; open challenges clarify what reliability still requires in deployment.
	As learned PDE solvers enter scientific workflows, predictability under contract shift, physical validity, algorithmic stability, and cost transparency become central concerns.
	
	\paragraph{Controlled extrapolation via explicit contracts.}
	Most evaluations emphasize interpolation, whereas deployment rarely does.
	Boundary conditions change, geometries vary, parameters cross regimes, and long rollouts amplify small biases.
	The most damaging failures are often silent.
	A central challenge is therefore \emph{controlled extrapolation}: specifying a generalization contract and instrumenting solvers with diagnostics.
	Proposal--correct--validate workflows are promising, where operators generate fast candidates (Eq.~\eqref{eq:operator_map}), and physics-based diagnostics determine whether to accept, refine, or reject them (Eq.~\eqref{eq:residual}).
	A concrete direction is to standardize stress tests along geometry, resolution, and rollout axes, aligned with the declared contract.
	
	\paragraph{Physics-first design as an interface.}
	Residual regularization improves robustness but is insufficient.
	PDEs encode conservation laws, symmetries, admissibility, and stability constraints.
	Violations can invalidate downstream decisions even when pointwise errors are small.
	An emerging opportunity is to treat physics as an explicit interface, with constraints enforced through architecture, parameterization, and projection; hard and semi-hard mechanisms enable correctness by construction~\cite{CHEN2021110624,chalapathi2024scalingphysicsinformedhardconstraints}.
	
	\paragraph{Training as a numerical method.}
	Complementarily, many limitations are algorithmic rather than representational.
	PINNs involve stiff, multi-objective optimization; neural operators learn nonlocal mappings sensitive to discretization, truncation, and data coverage.
	Both can fail due to poor conditioning.
	Accordingly, a more principled view treats training as a numerical process: sampling, resolution, and loss weighting play roles analogous to discretization and step-size selection in classical solvers across workflows.

	\paragraph{Scaling laws and end-to-end cost transparency.}
	Amortization motivates operator learning, but reported benefits can be incomplete.
	True cost is end-to-end: data generation, training, adaptation under shift, and debugging.
	The field lacks explicit cost models and scaling laws relating error to data volume, resolution, and compute across PDE families.
	Such transparency clarifies when operator learning outperforms reduced-order modeling, when physics-constrained learning outperforms adjoint-based inversion, and when hybrid approaches are necessary.
	Progress likely requires reporting a single cost ledger that includes simulation, training, fine-tuning, and deployment latency.

	\section{Conclusion}
	\label{sec:conclusion}
	
	This survey presented a unifying view of learning PDE solvers with physics and data.
	By placing PINN and NO within a shared, workflow-driven design space, we elucidate how differences in what is learned, how physics and data enter supervision, and where computation is amortized give rise to distinct strengths, limitations, and failure modes. 
	The significance of this unification lies in the growing reliance of modern scientific computing on reusable, data-integrated, and decision-driven PDE solvers that operate across heterogeneous workflows. 
	Progress in this area depends on moving towards shared abstractions, contracts, and evaluation signals that connect physics and data in a principled way.
	We hope that this survey can serve as a common reference that enables community-wide contributions towards the development of reliable learning-based PDE solvers for scientific discovery.

	\small
	\bibliographystyle{named}
	\bibliography{references_arxiv_backup.bib}
	
\end{document}